\documentclass[11pt]{article}
\usepackage{acl2015}
\usepackage{times}
\usepackage{url}
\usepackage{latexsym}
\usepackage{amsmath}
\usepackage{xcolor}
\usepackage{tikz}

\usepackage{algorithm}
\usepackage{algorithmic}

\title{Non-linear Learning for Statistical Machine Translation}

\author{First Author \\
  Affiliation / Address line 1 \\
  Affiliation / Address line 2 \\
  Affiliation / Address line 3 \\
  {\tt email@domain} \\\And
  Second Author \\
  Affiliation / Address line 1 \\
  Affiliation / Address line 2 \\
  Affiliation / Address line 3 \\
  {\tt email@domain} \\}

\date{}

\begin{document}
\maketitle
\begin{abstract}
Modern statistical machine translation (SMT) systems usually use a linear combination of features to model the quality of each translation hypothesis. The linear combination assumes that all the features are in a linear relationship and constrains that each feature interacts with the rest features in an linear manner, which might limit the expressive power of the model and lead to a under-fit model on the current data. In this paper, we propose a non-linear modeling for the quality of translation hypotheses based on neural networks, which allows more complex interaction between features. A learning framework is presented for training the non-linear models. We also discuss possible heuristics in designing the network structure which may improve the non-linear learning performance. Experimental results show that with the basic features of a hierarchical phrase-based machine translation system, our method produce translations that are better than a linear model.

\end{abstract}

\section{Introduction}
One of the core problems in the research of statistical machine translation is the modeling of translation hypotheses. Each modeling method defines a score of a target sentence $\mathbf{e}=e_1,e_2,...,e_i,...,e_I$, given a source sentence $\mathbf{f}=f_1,f_2,...,f_j...f_J$, where each $e_i$ is the $i$th target word and $f_j$ is the $j$th source word.
The well-known modeling method starts from the Source-Channel model~\cite{Brown1993}(Equation 1). The scoring of $\mathbf{e}$ decomposes to the calculation of a translation model and a language model.
\begin{equation}
Pr(\mathbf{e}|\mathbf{f})=Pr(\mathbf{e})Pr(\mathbf{f}|\mathbf{e})/Pr(\mathbf{f})
\end{equation} 

The modeling method is extended to log-linear models by Och and Ney~\shortcite{Och2002}, as shown in Equation 2, where $h_m(\mathbf{e}|\mathbf{f})$ is the $m$th feature function and $\lambda_m$ is the corresponding weight.
\begin{equation}
\begin{split}
&Pr(\mathbf{e}|\mathbf{f})=p_{\lambda_1^M}(\mathbf{e}|\mathbf{f})\\
&=\frac    {exp[\sum_{m=1}^M{\lambda_m h_m(\mathbf{e}|\mathbf{f})}]} 
 {{\sum_{\mathbf{e}'}  {exp[\sum_{m=1}^M{\lambda_m h_m(\mathbf{e}'|\mathbf{f})}]}  }} 
\end{split}
\end{equation} 

Because the normalization term in Equation 2 is the same for all translation hypotheses of the same source sentence, the score of each hypothesis, denoted by $s_L$, is actually a linear combination of all features, as shown in Equation \ref{eq-linear}. 
\begin{equation}\label{eq-linear}
s_{L}(\mathbf{e})=\sum_{m=1}^M{\lambda_m h_m(\mathbf{e}|\mathbf{f})}
\end{equation}

The log-linear models are flexible to incorporate new features and show significant advantage over the traditional source-channel models, thus become the state-of-the-art modeling method and are applied in various translation settings~\cite{Yamada2001,Koehn2003,Chiang2005,Liu2006}.

It is worth noticing that 
log-linear models try to separate good and bad translation hypotheses using a linear hyper-plane.
However, complex interactions between features make it difficult to linearly separate good translation hypotheses from bad ones~\cite{Clark2014}. 

Taking features in a typical phrase-based machine translation system~\cite{Koehn2003} as an example, the language model feature favors shorter hypotheses; the word penalty feature encourages longer hypotheses. The phrase translation probability feature selects phrases that occurs more frequently in the training corpus, which sometimes are long with lower translation probability, as in translating named entities or idioms; sometimes are short but with high translation probability, as in translating verbs or pronouns. These three features jointly decide the choice of translations. Simply use the weighted sum of their values may not be the best choice for modeling translations.

As a result, log-linear models may under-fit the data. This under-fitting may prevents the further improvement of translation quality. 



In this paper, we propose a non-linear modeling of translation hypotheses based on neural networks. The traditional features of a machine translation system are used as the input to the network. By feeding input features to nodes in a hidden layer, complex interactions among features are modeled, resulting in much stronger expressive power than traditional log-linear models. (Section~\ref{sec-translation})

Employing a neural network as non-linear models for SMT has two issues to be tackled. The first issue is the parameter learning. Log-linear models rely on minimum error rate training (MERT)~\cite{Och2003a} to achieve best performance. When the scoring function become non-linear, the intersection points of these non-linear functions could not be effectively calculated and enumerated. Thus MERT is no longer suitable for learning the parameters. To solve the problem , we present a framework for effective training including several criteria to transform the training problem into a binary classification task, a unified objective function and an iterative training algorithm. (Section~\ref{sec-framework})

The second issue is the structure of neural network. Single layer neural networks are equivalent to linear models; two-layer networks with sufficient nodes are capable of learning any continuous function~\cite{Bishop1995}. Adding more layers into the network could model complex functions with less nodes, but also brings the problem of vanishing gradient~\cite{Erhan2009}. We adapt a two-layer feed-forward neural network to keep the training process efficient. We notice that one major problem that prevent a neural network training reaching a good solution is that there are too many local minimums in the parameter space. Thus we discuss how to constrain the learning of neural networks with our intuition and observations of the features. (Section~\ref{sec-structure})

Experiments are conducted to compare various settings and verify the effectiveness of our proposed learning framework. Experimental results show that our framework could achieve better translation quality even with the same traditional features as previous linear models. (Section~\ref{sec-exp})



\section{Related work}


Many research has been attempting to bring non-linearity into the training of SMT. These efforts could be roughly divided into the following three categories.

The first line of research attempted to re-interpret original features via feature transformation or additional learning. For example, Maskey and Zhou~\shortcite{Maskey2012} use a deep belief network to learn representations of the phrase translation and lexical translation probability features. Clark et al.~\shortcite{Clark2014} used discretization to transform real-valued dense features into a set of binary indicator features. Lu et al.~\shortcite{Lu2014} learned new features using a semi-supervised deep auto encoder. These work focus on the explicit representation of the features and usually employ extra learning procedure. Our proposed method only take the original feature with no transformation as input. Feature transformation or combination are performed implicitly  during the training of the network and integrated with the optimization of translation quality.

The second line of research attempted to use non-linear models instead of log-linear models, which is most similar in spirit with our work.
Duh and Kirchhoff~\shortcite{Duh2008} used the boosting method to combine several results of MERT and achieved improvement in a re-ranking setting. Liu et al.~\shortcite{Liu2013} proposed an additive neural network which employed a two-layer neural network for embedding-based features. To avoid local minimum, they still rely on a pre-training and post-training from MERT or PRO. Comparing to these efforts, our proposed method takes a further step that it is integrated with iterative training, instead of re-ranking, and works without the help of any pre-trained linear models.

The third line of research attempted to add non-linear features/components into the log-linear learning framework. Neural network based models are trained as language models~\cite{Vaswani2013,Auli2014}, translation models~\cite{Gao2014} or joint language and translation models~\cite{Auli2013,Devlin2014}. Liu et al.~\shortcite{Liu2013} also introduced word embedding for source and target side of translation rule as local features. In this paper we focus on enhancing the expressive power of the modeling, which is independent of the research of enhancing translation system with new designed features. We believe additional improvement could be achieved by incorporating more features into our framework.

\section{Non-linear Translation}\label{sec-translation}
The non-linear modeling of translation hypotheses could be used in both phrase-based system and syntax-based systems. In this paper, we take the hierarchical phrase based machine translation system~\cite{Chiang2005} as an example and introduce how we fit the non-linearity into the system.

\subsection{Decoding}
The basic decoding algorithm could be kept almost the same as traditional phrase-based or syntax-based translation systems~\cite{Yamada2001,Koehn2003,Chiang2005,Liu2006}. For example, in the experiments of this paper, we use a CKY style decoding algorithm following Chiang~\shortcite{Chiang2005}. 

Our non-linear translation system is different from traditional systems in the way to calculate the score for each hypothesis. Instead of calculating the score as a linear combination, we use neural networks (Section \ref{ssec-nn}) to perform a non-linear combination of feature values.

We also use the cube-pruning algorithm~\cite{Chiang2005} to keep the decoding efficient. Although the non-linearity in model scores may cause more search errors in finding the highest scoring hypothesis, in practice it still achieves reasonable results. 

\subsection{Two-layer Neural Networks}\label{ssec-nn}
\begin{figure}
\begin{center}
\begin{tikzpicture}[x=16pt,y=36pt]
\foreach \x in {0,...,4}
  \node[draw,circle,shift={(-3.5,0)}](n\x{}2) at (\x,2) {};
\foreach \x in {0,...,5}
  \node[draw,circle,shift={(-4,0)}](n\x{}3) at (\x,3) {};
\foreach \x in {0,...,2}
  \node[draw,circle,shift={(-2.5,0)}](n\x{}4) at (\x,4) {};
\foreach \x in {0,...,4} \foreach \y in {0,...,5} \draw (n\x{}2.north) -- (n\y{}3.south);
\foreach \x in {0,...,5} \foreach \y in {0,...,2} \draw (n\x{}3.north) -- (n\y{}4.south);
\newcommand{\stack}[1]{\begin{tabular}{@{}c@{}}#1\end{tabular}}
\node at (-7,2) { \stack{input}};
\node at (-7,3) { \stack{hidden \\ layer}};
\node at (-7,4) { \stack{output \\ layer}};
\node at (-4,3.5) {$\mathbf{M_o}$};
\node at (-5,2.5) {$\mathbf{M_h}$};
\end{tikzpicture}
\end{center}
\caption{A two-layer feed-forward neural network.}
\label{fig-2layer}
\end{figure}
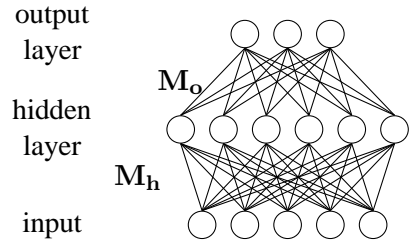

We employ a two-layer neural network as the non-linear model for scoring translation hypotheses. The structure of a typical two-layer feed-forward neural network includes an input layer,  a hidden layer, and a output layer (as shown in Figure \ref{fig-2layer}). 

We use the input layer to accept input features, the hidden layer to combine different input features, the output layer with only one node to output the model score for each translation hypothesis based on the value of hidden nodes. More specifically, the score of hypothesis $\mathbf{e}$, denoted as $s_N$, is defined as:
\begin{equation}\label{eq-nonlinear}
s_N(\mathbf{e})=\sigma_o(M_o\cdot\sigma_h(M_h\cdot h_1^m(\mathbf{e}|\mathbf{f})+b_h)+b_o)
\end{equation} 
where $M$, $b$ is the weight matrix, bias vector of the neural nodes, respectively; $\sigma$ is the activation function, which is often set to non-linear functions such as the tanh function or sigmoid function; subscript $h$ and $o$ indicates the parameters of hidden layer and output layer, respectively. 

\subsection{Features}\label{ssec-feature}
We use the standard features of a typical hierarchical phrase based translation system\cite{Chiang2005}. Adding new features into the framework is left as a future direction.
The features as listed as following:
\begin{itemize}
\item $p(\alpha|\gamma)$ and $p(\gamma|\alpha)$: conditional probability of translating $\alpha$ as $\gamma$ and translating $\alpha$ as $\gamma$, where $\alpha$ and $\gamma$ is the left and right hand side of a initial phrase (or hierarchical translation rule), respectively; 
\item $p_w(\alpha|\gamma)$ and $p_w(\gamma|\alpha)$: lexical probability of translating words in $\alpha$ as words in $\gamma$ and translating words in $\gamma$ as words in $\alpha$;
\item $p_{lm}$: language model probability;
\item $wc$: accumulated count of individual words generated during translation;
\item $pc$: accumulated count of initial phrases used;
\item $rc$: accumulated count of hierarchical rule phrases used;
\item $gc$: accumulated count of glue rule used in this hypothesis;
\item $uc$: accumulated count of unknown source word;
\item $nc$: accumulated count of source phrases that translate into null;
\end{itemize}

\section{Non-linear Learning Framework}\label{sec-framework}
Traditional machine translation systems rely on MERT to tune the weight of different features. MERT performs efficient search by enumerating the score function of all the hypotheses and using intersections of these linear functions to form the "upper-envelope" of the model score function~\cite{Och2003a}. When the scoring function is non-linear, it is not feasible to find the intersections of these functions. In this section, we discuss alternatives to train the parameter for non-linear models.

\subsection{Training Criteria}\label{ssec-intuion}

The task of machine translation is a complex problem with structural output space. Decoding algorithms search for the translation hypothesis with the highest score, according to a given scoring function, from an exponentially large set of candidate hypotheses. The purpose of training is to select the scoring function, so that the function score the hypotheses "correctly". The correctness is often introduced by some extrinsic metrics, such as BLEU~\cite{Papineni2002}.

We denote the scoring function as $s(\mathbf{f},\mathbf{e};\vec{\theta})$, or simply $s$, which is parametrized by $\vec{\theta}$; denote the set of all candidate hypotheses as $C$; denote the extrinsic metric as $eval(\cdot)$. Note that, in linear cases, $s$ is a linear function as in Equation \ref{eq-linear}, while in the non-linear case described in this paper, $s$ is the scoring function in Equation \ref{eq-nonlinear}.

Ideally, the training objective is to select a scoring function $s$, from all functions $\mathcal{S}$, that scores the correct translation (or references), denoted as $\hat{\mathbf{e}}$, higher than any other hypotheses (Equation \ref{eq-oracle}).
\begin{equation}\label{eq-oracle}
s = \{s\in \mathcal{S}|  s(\hat{\mathbf{e}})>s(\mathbf{e}) \, \forall \mathbf{e} \in C\}
\end{equation}

In practice, the candidate set $C$ is exponentially large and hard to enumerate; the correct translation $\hat{\mathbf{e}}$ may not even exist in the current search space for various reasons, e.g. unknown source word. As a result, we seek the following three alternatives as approximations to the ideal objective. 

\begin{description}
\item[Best v.s. Rest (BR)] To score the best hypothesis in the n-best set $\tilde{\mathbf{e}}$ higher than the rest hypotheses. This objective is very similar to MERT in that it tries to optimize the score of $\tilde{\mathbf{e}}$ and doesn't concern about the ranking of rest hypothesis. In this case, the n-best set $C_{nbest}$ is used to approximate $C$, and $\tilde{\mathbf{e}}$ to approximate $\hat{\mathbf{e}}$.

\item[Best v.s. Worst (BW)] To score the best hypothesis higher than the worst hypothesis in the n-best set. This objective is motivated by the practice of separating the "hope" and "fear" translation hypothesis \cite{Chiang2012}. We take a simpler strategy which uses the best and worst hypothesis in $C_{nbest}$ as the "hope" and "fear" hypothesis, respectively, in order to avoid multi-pass decoding.

\item[Pairwise (PW)] To score the better hypotheses in sampled hypothesis pairs higher than the worse ones in the same pair.
This objective is adapted from the Pairwise Ranking Optimization (PRO)~\cite{Hopkins2011}, which tries to ranking all the hypotheses instead of selecting the best one. We use the same sampling strategy as their original paper.
\end{description}

Note that each of the above criterions transforms the original problem of selecting best hypotheses from an exponential space to a certain pair-wise comparison problem, which could be easily trained as standard binary classifiers.

\subsection{Training Objective}\label{ssec-obj}
For the binary classification task, we use a hinge loss following Watanabe~\shortcite{Watanabe2012}. Because the network has a lot of parameters compared with the linear model, we use a $L_1$ norm instead of $L_2$ norm as the regularization term, to favor sparse solutions. We define our training objective function in Equation \ref{eq-obj}.
\begin{equation}\label{eq-obj}
\begin{split}
arg\min_\theta \, & \,  \frac{1}{N}\sum_{\mathbf{f}\in D}{\sum_{(\mathbf{e}_1,\mathbf{e}_2)\in T}   \delta(\mathbf{f}, \mathbf{e}_1, \mathbf{e}_2;\theta)}  +\lambda\cdot||\theta||_1 \\
\text{with} & \\
\delta(\cdot) &= max \{s(\mathbf{f},\mathbf{e}_1;\theta)-s(\mathbf{f},\mathbf{e}_2;\theta)+1, 0\}
\end{split}
\end{equation}
$D$ is the given training data; $(\mathbf{e}_1,\mathbf{e}_2)$ is a training hypothesis-pair, with the assumption that $\mathbf{e}_1$ is the one with higher $eval(\cdot)$ score; $N$ is the total number of hypothesis-pairs in $D$; $T$ is the set of hypothesis-pairs for each source sentence. 

The set $T$ is decided by the criterion used for training. For the BR setting, the best hypothesis is paired with every other hypothesis in the n-best list (Equation \ref{eq-br}); while for the BW setting, it is only paired with the worst hypothesis (Equation \ref{eq-bw}). The generation of $T$ in PW setting is the same with PRO sampling, we refer the readers to the original paper of Hopkins and May~\shortcite{Hopkins2011}.
\begin{equation}\label{eq-br}
\begin{split}
T_{BR}=\{(\mathbf{e}_1,\mathbf{e}_2)| & \mathbf{e}_1=arg\max_{\mathbf{e}\in C_{nbest}} eval(\mathbf{e}), \\
& \mathbf{e}_2\in C_{nbest} \,\text{and}\, \mathbf{e}_1 \neq \mathbf{e}_2\}
\end{split}
\end{equation}
\begin{equation}\label{eq-bw}
\begin{split}
T_{BW}=\{(\mathbf{e}_1,\mathbf{e}_2)| & \mathbf{e}_1=arg\max_{\mathbf{e}\in C_{nbest}} eval(\mathbf{e}), \\
 \mathbf{e}_2 & =arg\min_{\mathbf{e}\in C_{nbest}} eval(\mathbf{e})\}
\end{split}
\end{equation}

\subsection{Training Procedure}\label{ssec-tp}
In standard training algorithm for classification, the training instances stays the same in each iteration. In machine translation, decoding algorithms usually return a very different n-best set with different parameters. This is due to the exponentially large size of search space. MERT and PRO extend the current n-best set by merging the n-best set of all previous iterations into a pool~\cite{Papineni2002,Hopkins2011}. In this way, the enlarged n-best set may give a better approximation of the true hypothesis set $C$ and may lead to better and more stable training results. 

We argue that the training should still focus on hypotheses obtained in current round, because in each iteration the searching for the n-best set is independent of previous iterations. To compromise the above two goals, in our practice, training hypothesis pairs are first generated from the current n-best set, then merged with the pairs generated from all previous iterations. In order to make the model focus more on pairs from current iteration, we assign pairs in previous iterations a small constant weight and assign pairs in current iteration a relatively large constant weight. This is inspired by the AdaBoost algorithm \cite{Schapire1999} in weighting instances.

Following the spirit of MERT, we propose a iterative training procedure (Algorithm 1). 
\begin{algorithm}
\caption{Iterative Training Algorithm}
\label{alg-training}
\begin{algorithmic}[1]
\REQUIRE  the set of training sentences $D$, max number of iteration $I$
\STATE $\theta^0 \leftarrow \text{RandomInit}()$,
\FOR{ $i=0$ to $I$}
\STATE $T_{i} \leftarrow \emptyset$;
\FOR{ each $\mathbf{f} \in D$}
\STATE {$C_{nbest} \leftarrow  \text{NbestDecode} (\mathbf{f}$ ; $\theta^i)$} 
\STATE {$T \leftarrow \text{GeneratePair}(C_{nbest})$}
\STATE {$T_{i} \leftarrow T_{i}\cup T$}
\ENDFOR  
\STATE {$T_{all} \leftarrow \text{WeightedCombine}(\cup_{k=0}^{i-1}T_k, T_i)$}
\STATE { $\theta^{i+1} \leftarrow \text{Optimize} (T_{all}, \theta^i)$ }
\ENDFOR
\end{algorithmic}
\end{algorithm}

As shown in Algorithm 1, the training procedure starts by randomly init model parameters $\theta^0$ (line 1). In $i$th iteration, the decoding algorithm decodes each sentence $\mathbf{f}$ to get the n-best set $C_{nbest}$ (line 5). Training hypothesis pairs $T$ are extracted from $C_{nbest}$ according to the training criterion described in Section \ref{ssec-obj} (line 6). New collected pairs $T_i$ are combined with pairs from previous iterations before used for training (line 9). $\theta^{i+1}$ is obtained by solving Equation \ref{eq-obj} using the Conjugate Sub-Gradient method~\cite{Le2011} (line 10).

\section{Structure of the Network}\label{sec-structure}
Although neural networks bring strong expressive power to the modeling of translation hypothesis, training a neural network is prone to resulting in local minimum which may affect the training results. We speculate that one reason for these local minimums is the structure of a well-connected network has too many parameters. Take a neural network with $k$ nodes in the input layer and $m$ nodes in the hidden layer as an example. Every node in the hidden layer is connected to each of the $k$ input nodes. This simple structure resulting in at least $k\times m$ parameters. 

In Section \ref{ssec-obj}, we use $L_1$ norm in the objective function in order to get sparser solutions. In this section, we propose some constrained network structures according to our prior knowledge of the features. These structures have much less parameters or simpler structures comparing to original neural networks, thus reduce the possibility of getting stuck in local minimums.

\subsection{Network with two-degree Hidden Layer}
We find the first pitfall of the standard two-layer neural network is that each node in the hidden layer receives input from every input layer node. Features used in SMT are usually manually designed, which has their concrete meanings. For a network of several hidden nodes, combining every features into every hidden node may be redundant and not necessary to represent the quality of a hypothesis.

As a result, we take a harsh step and constrain the nodes in hidden layer to have a in-degree of two, which means each hidden node only accepts inputs from two input nodes. We do not use any other prior knowledge about features in this setting. So for a network with $k$ nodes in the input layer, the hidden layer should contain $C^2_k=k(k-1)/2$ nodes to accept all combinations from the input layer. We name this network structure as Two-Degree Hidden Layer Network (TDN).

It is easy to see that a TDN has $C^2_k \times 2 = k(k-1)$ parameters for the hidden layer because of the constrained degree. This is one order of magnitude less than a standard two-layer network with the same number of hidden nodes, which has $C^2_k \times k = k^2(k-1)/2$ parameters.

Note that we perform a 2-degree combination that looks similar in spirit with those combination of atomic features in large scale discriminative learning for other NLP tasks, such as POS tagging and parsing. However, unlike the practice in these tasks that directly combines values of different features to generate a new feature type, we first linearly combine the value of these features and perform non-linear transformation on these values via an activation function. 

\subsection{Network with Grouped Features}
It might be a too strong constraint to require the hidden node have in-degree of 2. In order to relax this constraint, we need more prior knowledge from the features. Our first observation is that there are different types of features. These types are different
from each other in terms of value ranges, sources, importance, etc. For example, language model features usually take a very small value of probability, and word count feature takes a integer value and usually has a much higher weight in linear case than other count features.
 
The second observation is that features in the same group are basically of the same type and may not have complex interaction with each other. For example, it is reasonable to combine language model features with word count features in a hidden node. But it may not be necessary to combine the count of initial phrases and the count of unknown words into a hidden node.

Based on the above two intuitions, we design a new structure of network that has the following constraints: given a disjoint partition of features: G$_1$, G$_1$, G$_k$, every hidden node takes input from a set of input nodes, where any two nodes in this set come from two different feature groups. We name this network structure as Grouped Network (GN).

In practice, we divide the basic features in Section~\ref{ssec-feature} into five groups: language model features, translation probability features, lexical probability features, the word count feature, and the rest of count features.

\section{Experiments and Results}\label{sec-exp}
\subsection{General Settings}
We conduct experiments on a large scale machine translation tasks. The parallel data comes from LDC, including LDC2002E18, LDC2003E14, LDC2004E12, LDC2004T08, LDC2005T10, LDC2007T09, which consists of 8.6 million of sentence pairs. Monolingual data includes Xinhua portion of Gigaword corpus. We use multi-references data MT03 as training data, MT02 as development data, and MT04, MT05 as test data. These data are mainly in the same genre, avoiding the extra consideration of domain adaptation.

\begin{table}[ht]
\centering
\begin{tabular}{c|c|cc}
\cline{1-3}
  Data  & Usage & Sents. &  \\ \cline{1-3}
  LDC  & TM train & 8,260,093 &  \\ \cline{1-3}
  Gigaword  & LM train& 14,684,074 &  \\ \cline{1-3}
  MT03 & train & 919 & \\ \cline{1-3}
  MT02   & dev & 878 & \\ \cline{1-3}
  MT04   & test & 1,789 & \\ 
  MT05   & test & 1,083& \\ \cline{1-3}
\end{tabular}
\caption{Experimental data and statistics.}
\label{tb-data}
\end{table}

The Chinese side of the corpora is word segmented using ICTCLAS\footnote{http://ictclas.nlpir.org/}. Our translation system is an in-house implementation of the hierarchical phrase-based translation system\cite{Chiang2005}. We set the beam size to 20. We train a 5-gram language model on the monolingual data with MKN smoothing\cite{Chen1998}. For each parameter tuning experiments, we ran the same training procedure 3 times and present the average results. The translation quality is evaluated use 4-gram case-insensitive BLEU~\cite{Papineni2002}. Significant test is performed using bootstrap re-sampling implemented by Clark et al.~\shortcite{Clark2011}.
\begin{table*}[ht]
\centering
\begin{tabular}{c|c|c|c|cc}
\cline{1-5}
  Criteria  & MT03(train) & MT02(dev) & $\quad$MT04$\quad$ & $\quad$MT05$\quad$ & \\ \cline{1-5}
  BR$_c$  & 35.02 & 36.63 & 34.96 & 34.15 &  \\   
  BR  & 38.66 & 40.04 & 38.73 & 37.50 &  \\  
  BW & 39.55 & 39.36 & 38.72 & 37.81 &  \\    
  PW & 38.61 & 38.85 & 38.73 & 37.98 & \\ \cline{1-5}
\end{tabular}
\caption{BLEU4 in percentage on different training criteria ("BR", "BW" and "PW" refer to experiments with "Best v.s. Rest", "Best v.s. Worst" and "Pairwise" training criteria, respectively. "BR$_c$" indicates generate hypothesis pairs from n-best set of current iteration only presented in Section~\ref{ssec-tp}.}
\label{tb-criteria}
\end{table*}
We employ a two-layer neural network with 11 input layer nodes, corresponding to features listed in Section~\ref{ssec-feature} and 1 output layer node. The number of nodes in the hidden layer varies in different settings. The sigmoid function is used as the activation function for each node in the hidden layer. For the output layer we use a linear activation function. We try different $\lambda$ for the L$_1$ norm from 0.01 to 0.00001 and use the one with best performance on the development set. We solve the optimization problem with ALGLIB package\footnote{http://www.alglib.net/}. 

\subsection{Experiments of Training Criteria}
This set experiments evaluates different training criteria discussed in Section~\ref{ssec-intuion}. We generate hypothesis-pair according to BW, BR and PW criteria, respectively, and perform training with these pairs. In the PW criterion, we use the sampling method of PRO~\cite{Hopkins2011} and get the 50 hypothesis pairs for each sentence. We use 20 hidden nodes for all three settings to make a fair comparison. 

The results are presented in Table~\ref{tb-criteria}. The first two rows compare training with and without the weighted combination of hypothesis pairs we discussed in Section~\ref{ssec-tp}. As the result suggested, with the weighted combination of hypothesis pairs from previous iterations, the performance improves significantly on both test sets.

Although the system performance on the dev set varies, the performance on test sets are almost comparable. This suggest that although the three training criteria are based on different assumptions, their are basically equivalent for training translation systems. 

\begin{table}[th]
\centering
\begin{tabular}{c|c|cc}
\cline{1-3}
  Criteria  & Pairs/iteration & Accuracy(\%)   & \\ \cline{1-3}
  BR  & 19 & 70.7 &   \\  
  BW & 1 & 79.5 &    \\    
  PW & 100 & 67.3 &   \\ \cline{1-3}
\end{tabular}
\caption{Comparison of different training criteria in number of new instances per iteration and training accuracy.}
\label{tb-ct2}
\end{table}

We also compares the three training criteria in their number of new instances per iteration and final training accuracy (Table~\ref{tb-ct2}). Compared to BR which tries to separate the best hypothesis from the rest hypotheses in the n-best set, and PW which tries to obtain a correct ranking of all hypotheses, BW only aims at separating the best and worst hypothesis of each iteration, which is a easier task for learning a classifiers. It requires the least training instances and achieves the best performance in training. Note that, the accuracy for each system in Table~\ref{tb-ct2} are the accuracy each system achieves after training stops. They are not calculated on the same set of instances, thus not directly comparable. We use the differences in accuracy as an indicator for the difficulties of the corresponding learning task.

For the rest of this paper, we use the BW criterion because it is much simpler compared to sampling method of PRO~\cite{Hopkins2011}.

\begin{table*}[ht]
\centering
\begin{tabular}{c|c|c|c|c|cc}
\cline{1-6}
  Systems  & MT03(train) & MT02(dev) & $\quad$MT04$\quad$ & $\quad$MT05$\quad$ & TestAverage &\\ \cline{1-6}
  HPB  & 39.25$^+$&	39.07&	38.81&	38.01&  38.41(-)&\\    \cline{1-6}
  TLayer$_{20}$ & 39.55$^*$ & 39.36$^*$ & 38.72 & 37.81 & 38.27(-0.14) \\    
  TLayer$_{30}$ &39.70$^+$&	39.71$^*$&	38.89&	37.90& 38.40(-0.01)& \\  
  TLayer$_{50}$ &39.26&	38.97&	38.72&	38.79$^+$& 38.76(+0.35)& \\ \cline{1-6}
  TDN & 39.60$^+$&	38.94&	38.99$^*$&	38.13& 38.56(+0.15)& \\    
  GN & \textbf{39.73}$^+$&	\textbf{39.41}$^+$&	\textbf{39.45}$^+$&	\textbf{38.51}$^+$& \textbf{38.98(+0.57)}&\\ \cline{1-6}
\end{tabular}
\caption{BLEU4 in percentage for comparing of systems using different network structures (HPB refers to the baseline hierarchical phrase-based system. TLayer , TDN, GN refer to the standard 2-layer network, Two-Degree Hidden Layer Network, Grouped Network, respectively. Subscript of TLayer indicates the number of nodes in the hidden layer.) $^+$, $^*$ marks results that are significant better than the baseline system with $p<0.01$ and $p<0.05$.}
\label{tb-systems}
\end{table*}

\subsection{Experiments of Network Structures}
We make several comparisons of the network structures and compare them with a baseline hierarchical phrase-based translation system (HPB) (Table~\ref{tb-systems}). 

We first compares the neural network with different number of hidden nodes. The systems TLayer$_{20}$, TLayer$_{30}$ and TLayer$_{50}$ are standard two-layer feed forward neural network with 20, 30 and 50 hidden layer nodes\footnote{TLayer$_{20}$ is the same system as BW in Table~\ref{tb-criteria}}. We can see that training a larger network do lead to an improvement in translation quality. However training a larger network is often time-consuming. We experimented with neural networks with 100 and more hidden nodes (TLayer$_{100}$ ). But TLayer$_{30}$  takes 10 times longer in training time for each iteration than TLayer$_{20}$ and did not finish by the time of submission deadline.

We then compared the two network structures proposed in Section~\ref{sec-structure}. The Two-Degree Hidden Layer Network (TDN) already perform comparable to the baseline system. But it constrain all input to the hidden node to be of degree 2, which is likely to be too restrictive. With the grouped feature, we could design networks such as GN, which shows significant improvement over the baseline systems and achieves the best performance among all neural systems. Note that GN is in a much larger scale, but is also sparse in parameters and takes significant less training time than standard neural networks.

\section{Conclusion}
In this paper, we discuss a non-linear framework for modeling translation hypothesis for statistical machine translation system. We also present a learning framework including training criterion and algorithms to integrate our modeling into a state of the art hierarchical phrase based machine translation system.  Compared to previous effort in bringing in non-linearity into machine translation, our method uses a single two-layer neural networks and performs training independent with any previous linear training methods (e.g. MERT). Our method also trains its parameters without any pre-training or post-training procedure. Experiment shows that our method could improve the baseline system even with the same feature as input, in a large scale Chinese-English machine translation task.

In training neural networks with hidden nodes, we use heuristics to reduce the complexity of network structures and obtain extra advantages over standard networks. It shows that heuristics and intuitions of the data and features are still important to a machine translation system. 

As future work, it is necessary to integrate more features into our learning framework.  It is also interesting to see how the non-linear modeling fit in to more complex learning tasks which involves domain specific learning techniques.


\bibliographystyle{acl}
\bibliography{SMT}

\end{document}